\documentclass[a4paper, 10pt, conference]{ieeeconf}      

\IEEEoverridecommandlockouts                              
\usepackage{tikz}
\usepackage{subcaption}
\usepackage{amsmath,amssymb}
\usepackage{algorithm}
\usepackage{algpseudocode}
\usepackage{graphicx}
\usepackage{booktabs}
\usepackage{url}
\usepackage{xcolor}
\usepackage{hyperref}
\usepackage[table]{xcolor}
\definecolor{ourrow}{gray}{0.94}

\usepackage{booktabs}
\usepackage{textcomp}
\usepackage{mdframed}
\usepackage{censor}                                         
\usepackage{booktabs}
\usepackage{multirow}
\title{\LARGE \bf
Speculative Policy Orchestration: A Latency-Resilient Framework for Cloud-Robotic Manipulation
}

\author{
Chanh Nguyen$^{*1}$\thanks{*These authors contributed equally to this work.}, 
Shutong Jin$^{*2}$, 
Florian T. Pokorny$^{2}$,
Erik Elmroth$^{1}$%
\thanks{$^{1}$Chanh Nguyen and Erik Elmroth are with the Department of Computing Science,
Ume{\aa} University, 90187 Ume{\aa}, Sweden
        {\tt\small \{chanh, elmroth\}@cs.umu.se}}%
\thanks{$^{2}$Shutong Jin and Florian T. Pokorny are with School of Electrical Engineering and Computer Science,
KTH Royal Institute of Technology, 10044 Stockholm, Sweden
        {\tt\small \{shutong, fpokorny\}@kth.se}}%
}

\begin{document}

\maketitle
\thispagestyle{empty}
\pagestyle{empty}

\begin{abstract}
Cloud robotics enables robots to offload high-dimensional motion planning and reasoning to remote servers. However, for continuous manipulation tasks requiring high-frequency control, network latency and jitter can severely destabilize the system, causing command starvation and unsafe physical execution.

To address this, we propose Speculative Policy Orchestration (SPO), a latency-resilient cloud-edge framework. SPO utilizes a cloud-hosted world model to pre-compute and stream future kinematic waypoints to a local edge buffer, decoupling execution frequency from network round-trip time. To mitigate unsafe execution caused by predictive drift, the edge node employs an $\epsilon$-tube verifier that strictly bounds kinematic execution errors. The framework is coupled with an Adaptive Horizon Scaling mechanism that dynamically expands or shrinks the speculative pre-fetch depth based on real-time tracking error. 

We evaluate SPO on continuous RLBench manipulation tasks under emulated network delays. Results show that even when deployed with learned models of modest accuracy, SPO reduces network-induced idle time by over 60\% compared to blocking remote inference. Furthermore, SPO discards approximately 60\% fewer cloud predictions than static caching baselines. Ultimately, SPO enables fluid, real-time cloud-robotic control while maintaining bounded physical safety.

\end{abstract}

\section{INTRODUCTION}
\label{intro}

The integration of deep neural networks~\cite{xian2023chaineddiffuser, sunarch, gao2024prime} into robotic manipulation has enabled agents to perform increasingly complex, long-horizon tasks, shifting from rigid pre-programmed cycles toward flexible, embodied intelligence. While earlier paradigms abstracted these tasks into discrete semantic primitives (e.g., ``reach" or ``grasp")~\cite{gao2024prime}, state-of-the-art visuomotor policies increasingly rely on \textit{continuous action chunking}, i.e., predicting dense, high-frequency sequences of kinematic targets to achieve fine-grained dexterity~\cite{zhao2023learning, chi2025diffusion}.

Despite the algorithmic success of continuous, multi-task models like Hierarchical Diffusion Policy (HDP)~\cite{ma2024hierarchical} and Sequential Dexterity~\cite{chensequential}, these systems implicitly assume the availability of unconstrained, low-latency local computation capable of executing large Transformer and Diffusion-based networks in real-time. Scaling such architectures to industrial deployments motivates a cloud robotics paradigm, wherein heavy policy networks and predictive world models are offloaded to remote servers, allowing robots to leverage elastic computational resources.

Existing cloud robotics frameworks~\cite{brohan2023can, tian2019fog, ahn2024autort} primarily support high-level Vision–Language–Action (VLA) models by transmitting discrete \emph{macro-goals} (e.g., ``pick up the apple'' or ``move to the red bin''), where network delays are negligible relative to task-level timescales. In contrast, modern visuomotor policies require \textit{continuous kinematic waypoints} to be updated at high frequencies (e.g., $10-50$ Hz)~\cite{zhao2023learning, tu2019robust} to achieve smooth, dexterous motion. While the lowest-level hardware torque servoing (e.g., $500-1000$ Hz)~\cite{englsberger2014overview, wensing2017proprioceptive} safely remains on the robot's embedded controller, offloading the $10-50$ Hz kinematic trajectory generation to the cloud exposes a fundamental systems constraint: \textbf{network latency}~\cite{black2025real, chen2024fogros2}.

In real-world 5G or Wi-Fi deployments, link handovers and network congestion can cause delays ranging from hundreds of milliseconds to multiple seconds~\cite{liu2024m2ho, xu2020understanding}. When the network round-trip time exceeds the 20--100 ms control interval required for waypoint updates, the local torque controller experiences target starvation, resulting in discontinuous command updates, degraded motion smoothness, and accumulated tracking error. In contact-rich manipulation tasks, such disruptions can significantly compromise execution stability and task performance.

In large language models and cloud gaming systems, communication and computation latency are mitigated through speculative techniques that proactively predict and execute future states, such as \textit{speculative decoding}~\cite{leviathan2023fast, xu2024edgellm, li2024nearest,kim2023speculative} in autoregressive models and \textit{speculative execution} in computer architecture~\cite{lee2015outatime}. Inspired by these latency-masking mechanisms, we introduce \textbf{S}peculative \textbf{P}olicy \textbf{O}rchestration (\textbf{SPO}), a cloud-edge framework that bridges the latency gap in high-frequency continuous robotic control. Rather than proposing a new policy architecture, SPO operates as a \emph{model-agnostic orchestration layer} that coordinates cloud-hosted policies with edge-side execution.
Instead of waiting for a completed physical motion to transmit a new observation, the remote cloud utilizes a forward world model to speculatively generate a dense trajectory of the next $K$ kinematic waypoints. These state-action rollouts are streamed to the edge and buffered locally, thereby decoupling the kinematic planning frequency from network round-trip time while preserving a continuous Markov Decision Process (MDP) formulation. Crucially, low-level torque servoing remains strictly local to the robot’s embedded controller, ensuring real-time stability and preventing network-induced starvation.

\begin{figure*}[]
      \centering
      
      \includegraphics[width=0.7\textwidth]{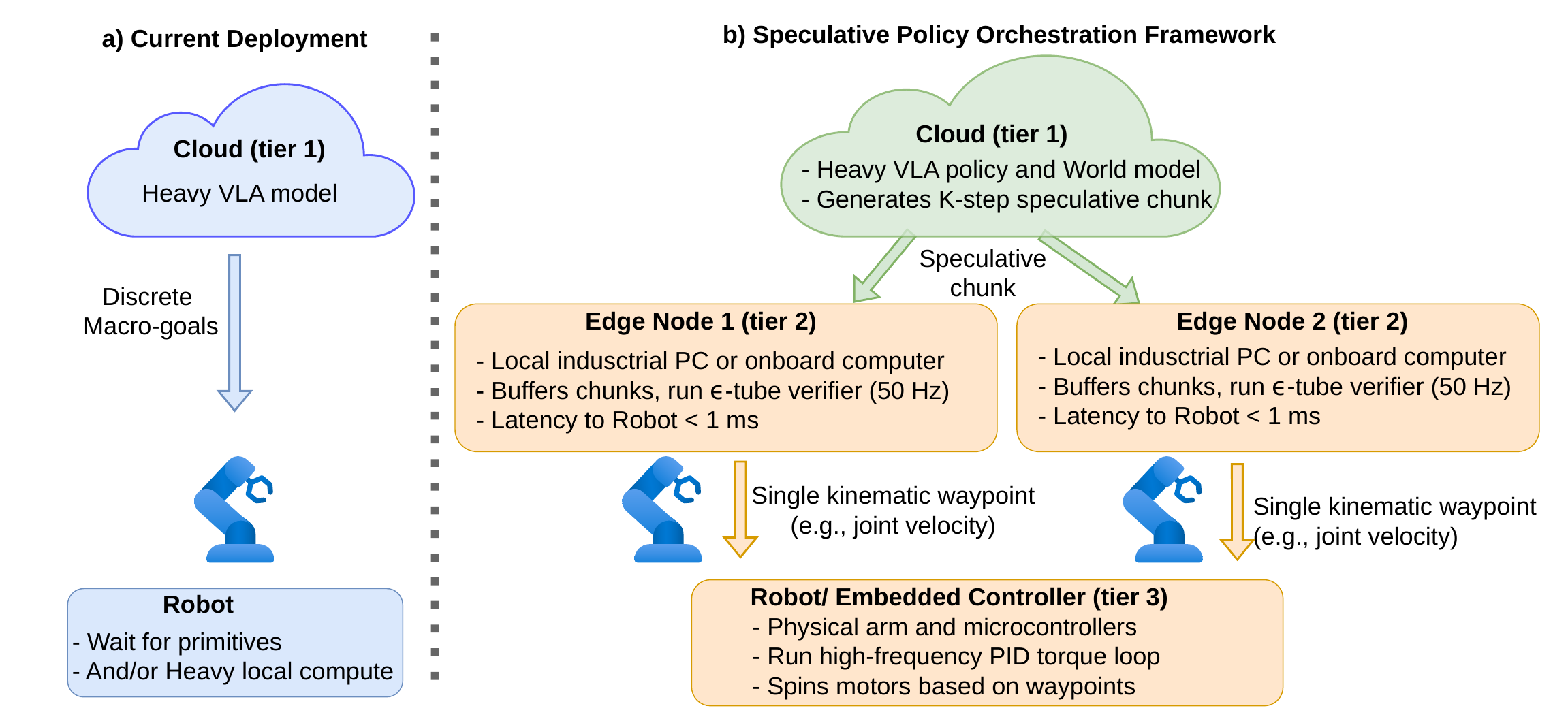}
    \caption{Comparison of traditional cloud robotics architectures with the proposed SPO framework. (a) Traditional deployments are bottlenecked by network latency, forcing them to rely on discrete macro-goals or expensive local compute. (b) SPO leverages a centralized World Model to proactively serve high-frequency, continuous kinematic chunks to a fleet of edge robots, enforcing kinematic safety limits via local $\epsilon$-tube verifiers. }
      \label{fig:overview}
   \end{figure*}

To ensure safe execution under speculative rollouts, SPO does not rely on potentially miscalibrated probabilistic confidence estimates from the neural policy. Instead, it enforces safety via an empirically defined kinematic error envelope, formalized as an $\epsilon$-tube around the predicted trajectory. If the observed state deviates beyond this bound, speculative execution is invalidated and replanning is triggered. Furthermore, SPO introduces an Adaptive Horizon Scaling (AHS) feedback mechanism that dynamically adjusts the speculative horizon depth based on real-time tracking error, allocating cache capacity proportionally to physical uncertainty. Figure~\ref{fig:overview} depicts the SPO framework and its advantages over traditional macro-goal cloud robotics deployments.

The key contributions of this paper are as follows:

\begin{itemize}
    \item \textbf{A Continuous Speculative Framework}: We introduce SPO, a \textit{model-agnostic cloud-edge architecture} that masks network latency for high-frequency continuous robotic control without relying on discrete behavior primitives (Section~\ref{spo}).
    
    \item \textbf{Kinematic Stability Bounds}: We formulate an edge-based $\epsilon$-tube verification mechanism that mitigates unsafe speculative execution by bounding kinematic tracking errors. The mechanism includes a zero-velocity \textit{hold} reflex to suspend unverified execution upon verification failure, bypassing the risks of miscalibrated neural probabilities (Section~\ref{e-tube}).
    
    \item \textbf{Adaptive Horizon Scaling}: We propose an AHS logic that dynamically scales the pre-fetch horizon based on real-time tracking errors, allows the system to scale bandwidth and cache memory linearly ($\mathcal{O}(K)$) rather than exponentially, adapting to physical uncertainty in real-time (Section~\ref{ahs}).
    
    \item \textbf{Empirical Validation}: We conduct extensive evaluations on the RLBench benchmark~\cite{james2020rlbench} using a physically distributed edge-cloud architecture subjected to an emulated network delay of 150 ms with a stochastic variation of 30 ms. Over tasks of varying complexity (\textit{StackBlocks}, \textit{InsertOntoSquarePeg}, \textit{PutAllGroceriesInCupboard}), SPO reduces network-induced idle time by over 60\% and optimizes resource utilization by decreasing wasted cloud predictions by 60\% relative to static caching baselines, demonstrating robust execution even with learned models of modest accuracy (Section~\ref{experiment}).
    
\end{itemize}

\section{RELATED WORK}

\subsection{Cloud Robotics Platforms} 
The paradigm of cloud robotics has emerged to address the computational limitations of onboard robotic platforms by offloading resource-intensive tasks to remote servers. Frameworks such as DAvinCi~\cite{arumugam2010davinci}, RoboEarth~\cite{waibel2011roboearth}, Rapyuta~\cite{mohanarajah2014rapyuta}, FogROS2~\cite{ichnowski2023fogros2}, KubeROS~\cite{zhang2023kuberos}, tinyKube~\cite{nguyen2025tinykube} provide communication and orchestration layers between robots at the edge and cloud infrastructure. However, these platforms primarily target perception and planning workloads and do not explicitly address the challenges of latency-sensitive closed-loop control required for continuous manipulation.

\subsection{Policy Learning and Sequence Prediction}
To manage complex manipulation tasks, early robot learning approaches relied on predefined motion primitives to reduce the dimensionality of continuous action spaces~\cite{gao2024prime, Chen_2025_CVPR,  xu2021efficient, tang2021learning, nasiriany2022augmenting,zhu2022bottom}. More recently, generative models such as Diffusion Policies \cite{wang2022diffusion, janner2022diffuser, reuss2023goal} predict continuous action sequences directly from observations. Similarly, Action Chunking with Transformers~\cite{zhao2023learning} and MT-ACT \cite{bharadhwaj2024roboagent} generate fixed-length macro-actions to improve temporal consistency and mitigate covariate shift.

However, deploying these multi-step predictive models over a cloud-edge continuum exposes a critical systems-level vulnerability. Once an action sequence is dispatched over the network, its execution at the edge becomes effectively open-loop.
As reported in MT-ACT~\cite{bharadhwaj2024roboagent}, executing large static chunks (e.g., $H=40$ steps) causes significant performance degradation due to the inability of the policy to correct compounding predictive errors once the chunk is dispatched.

Our proposed SPO framework addresses this limitation by operating beneath these predictive models as a \textit{model-agnostic orchestration layer}. Rather than treating sequences as rigid commands, SPO treats them as a speculative cache. By utilizing a local $\epsilon$-tube verifier and adaptive horizon scaling, SPO safely bounds execution errors and masks network latency, dynamically collapsing the cache at the edge only when physical verification fails.

\subsection{Latency-Resilient Control and Runtime Verification}

The concept of predicting future trajectories and enforcing bounded tracking errors around nominal paths can also be observed in receding-horizon control frameworks such as Model Predictive Control (MPC)~\cite{kouvaritakis2016model} and its robust variants (e.g., tube MPC)~\cite{mayne2005robust, rakovic2005invariant}. These methods typically rely on explicit system models and solve an optimization problem at each control step. 
In contrast, SPO does not perform online trajectory optimization. Instead, it verifies cloud-generated policy rollouts at the edge and dynamically adjusts the speculative horizon based on observed tracking error.

Recent efforts also explore compressing or distilling large visuomotor policies to enable low-latency local execution~\cite{prasad2024consistency,xu2024rldg, wang2024one}. SPO is orthogonal to these approaches: when policies can run locally, the speculative horizon naturally shrinks; when large policies remain cloud-hosted, SPO provides a latency-resilient execution layer that maintains continuous control despite network delays.

\section{PROBLEM FORMULATION}
\label{formulation}

\subsection{Continuous Manipulation as a Markov Decision Process}
We consider a discrete-time robotic manipulation task modeled as a deterministic Markov Decision Process (MDP), defined by the tuple $\langle \mathcal{S}, \mathcal{A}, \mathcal{T} \rangle$, where:
\begin{itemize}
    \item \textbf{State space} $\mathcal{S} \subset \mathbb{R}^{d_s}$: The continuous state $s_t$ represents the robot’s joint configuration, velocities, and task-relevant object poses.
    
    \item \textbf{Action space} $\mathcal{A} \subset \mathbb{R}^{d_a}$: The action $a_t$ corresponds to dense kinematic targets at the trajectory-generation layer (e.g., desired joint velocities $\dot{q}_t$).

    \item \textbf{Transition dynamics} $\mathcal{T}$: The system evolves according to $s_{t+1} = \mathcal{T}(s_t, a_t)$, where $\mathcal{T}$ captures the underlying physical dynamics.
\end{itemize}

Here, $d_s$ and $d_a$ denote the dimensionality of the state and action vectors, respectively. At each time step $t$, the robot observes $s_t$ and applies control input $a_t$, resulting in a new state $s_{t+1}$.


\subsection{Cloud-Induced Delay and Open-Loop Vulnerability}
In a standard cloud-based hierarchical control paradigm, the policy $\pi: \mathcal{S} \rightarrow \mathcal{A}$ is executed remotely. Consequently, the action selected at time $t$ is applied after a network round-trip delay $\tau_{\text{net}} > 0$, inducing a delayed control system $a_t = \pi(s_{t-\delta})$ for $t \ge \delta$, where $\delta = \left\lfloor \tau_{\text{net}} / \Delta t \right\rfloor$ and $\Delta t$ denotes the local control interval. 

When $\tau_{\text{net}} \gg \Delta t$, the system operates under delayed feedback: actions are computed from stale state information $s_{t-\delta}$ rather than the current state $s_t$. For high-frequency manipulation (e.g., $\Delta t = 10{-}20$ ms), even moderate network delays ($\tau_{\text{net}} = 30{-}100$ ms) produce $\delta > 1$, degrading closed-loop responsiveness and increasing tracking error.

To mask this delay, existing remote deployments often transmit action chunks, i.e., finite sequences $\{a_t, \dots, a_{t+K}\}$ for local buffered execution~\cite{zhao2023learning, chi2025diffusion}. While chunking alleviates command starvation, it transforms the system into a partially open-loop controller during chunk playback~\cite{bharadhwaj2024roboagent}. If unmodeled disturbances or contact events occur mid-chunk, the locally buffered trajectory may no longer be physically consistent with the true state evolution.
Blindly executing this stale trajectory through unanticipated collisions can lead to dangerous force accumulations and task failure.

Furthermore, fixing the speculative horizon $K$ introduces a fundamental tradeoff, i.e., larger horizons improve latency masking but amplify the impact of model error, whereas shorter horizons reduce open-loop exposure but reintroduce network blocking. Consequently, horizon selection must adapt to evolving physical uncertainty.

In the following section, we introduce SPO, a principled cloud-edge framework that compensates for network-induced delay while ensuring real-time physical validity and adaptive horizon regulation under dynamic uncertainty.

\section{SPECULATIVE POLICY ORCHESTRATION}
\label{spo}

   \begin{figure}
      \centering
      
      \includegraphics[width=0.8\linewidth]{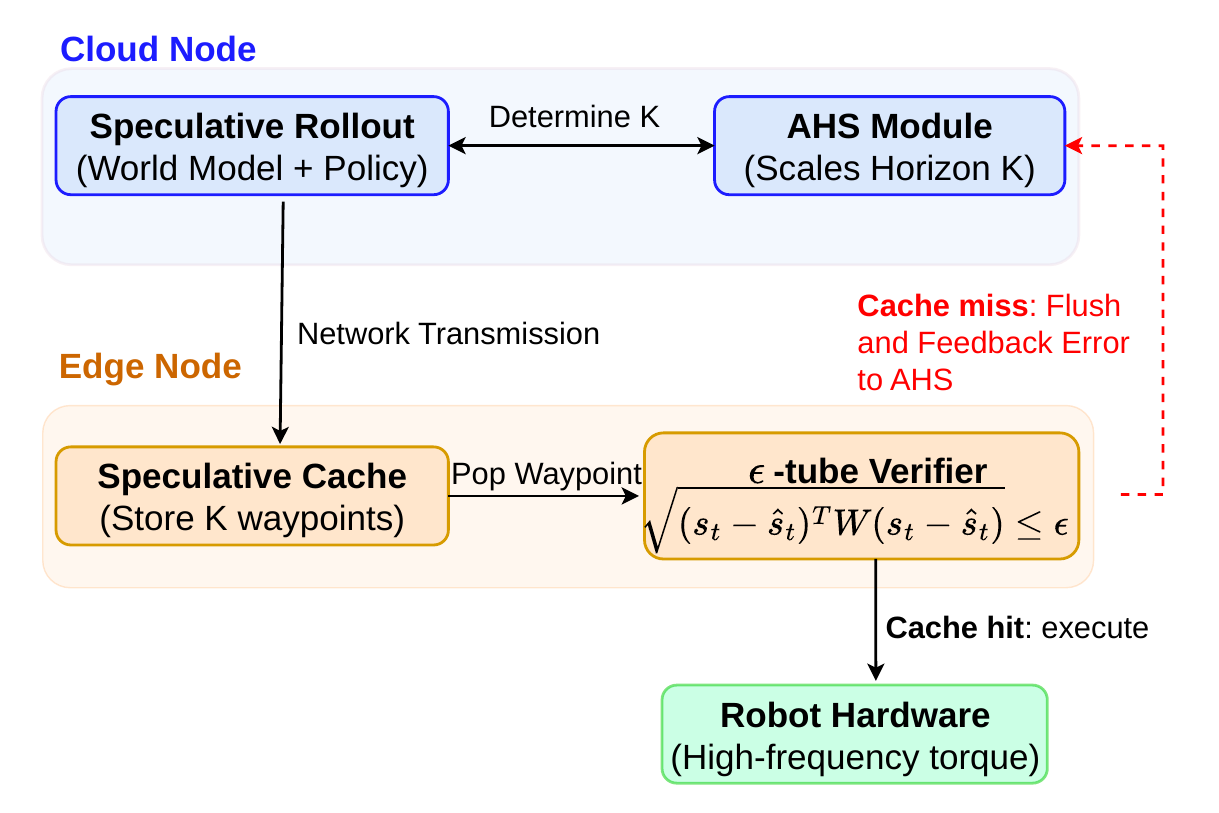}
    \caption{Architecture of SPO. The cloud executes computationally intensive speculative rollouts, adaptively scaled by the AHS module, while the edge enforces safety constraints using a lightweight $\epsilon$-tube verifier.}
      \label{fig:architecture_spo}
   \end{figure}

Figure~\ref{fig:architecture_spo} presents the architecture of SPO, a cloud–edge design that decouples high-dimensional trajectory generation from millisecond-scale physical actuation. The system consists of a remote Cloud World Model that performs speculative state–action rollouts and a local Edge Verifier that enforces safety and executes validated commands in real time.

Algorithm~\ref{alg:spo} outlines the system-level execution flow of SPO. Note that while the logic is presented chronologically within a single temporal control loop over time $t$, the execution is physically distributed. The Kinematic Verification phase (lines 5--19) evaluates strictly on the local Edge Node at high frequency. If a safety violation occurs, the edge node safely blocks, and the sequence transfers to the remote Cloud Server, which executes the AHS and Speculative Generation phase (lines 20--32) before transmitting the payload back to the edge to resume the loop.

\subsection{Cloud-Side Autoregressive Generation}
As defined in Section~\ref{formulation}, the environment evolves according to the true transition dynamics $s_{t+1} = \mathcal{T}(s_t, a_t)$. To mask the network latency $\tau_{\text{net}}$ associated with remote offloading, the Cloud Server approximates these dynamics using a learned World Model $\mathcal{M}_\phi$ and a high-frequency policy $\pi_\theta$.

Instead of waiting for the physical transition $\mathcal{T}$ to complete and transmit a new observation, the cloud generates a speculative continuous trajectory of the next $K$ kinematic steps. As shown in Algorithm~\ref{alg:spo} (lines 28--32), this is achieved via an autoregressive loop where the world model provides the necessary state feedback for subsequent policy queries:

$$\hat{a}_k = \pi_\theta(\hat{s}_k), \quad \hat{s}_{k+1} = \mathcal{M}_\phi(\hat{s}_k, \hat{a}_k)$$

for $k = 1, \dots, K$, where $\hat{s}_1$ is initialized with the most recent physical observation $s_t$. Each generated tuple $(\hat{s}_{k+1}, \hat{a}_k)$ is appended to the trajectory cache $\mathcal{C}$, thereby decoupling kinematic planning frequency from network-induced observation delay.

\subsection{Adaptive Horizon Scaling (AHS)}
\label{ahs}

A critical challenge in speculative orchestration is determining the optimal pre-fetch horizon ($K$). Static pre-fetching methods that cache deep, multi-second velocity plans without high-frequency validation risk catastrophic instability upon physical contact. Conversely, caching too few steps fails to mask network latency, leading to control starvation.

To balance latency masking and execution safety, SPO replaces static caching with an AHS mechanism that dynamically adjusts the speculative horizon based on real-time tracking errors. AHS acts as a lightweight feedback controller for cloud-side speculation, adapting the amount of pre-fetched trajectory to current physical conditions.

As defined in Algorithm~\ref{alg:spo} (lines 20--25), AHS follows an Additive-Increase/Multiplicative-Decrease rule:
\begin{itemize}
    \item \textbf{Free-Space Expansion:} When tracking remains within the verification bound ($e_t \le \epsilon_{base}$), the horizon expands gradually
    $K \leftarrow \min(K_{max}, K + \beta)$, where $\beta$ is the additive horizon increment,
    allowing the system to progressively increase latency masking during stable execution.

    \item \textbf{Contact-Induced Contraction:} When a divergence is detected ($e_{miss} > 0$), the horizon contracts according to
    $K \leftarrow \max(K_{min}, \lfloor K / \rho \rfloor)$,
    where $\rho = e_{miss} / \epsilon_{base}$. This limits speculative execution to short trajectory segments during contact-rich or highly dynamic phases.
\end{itemize}

By construction, the horizon remains bounded within $K \in [K_{min}, K_{max}]$, preventing unbounded algorithmic growth. Under sustained cache hits, $K$ increases linearly and reaches $K_{max}$ in at most $(K_{max}-K_{min})/\beta$ control cycles, while verification failures trigger multiplicative contraction proportional to the observed error severity $\rho$. This asymmetric expansion-contraction behavior allows the speculative horizon to adapt to changes in tracking error while limiting excessive oscillations under transient disturbances such as contact events or network jitter. Short-lived disturbances may temporarily reduce $K$, but the additive expansion rule allows the horizon to recover gradually once nominal tracking resumes.

\subsection{Continuous-State $\epsilon$-Tube Verification}
\label{e-tube}
Executing an incorrect speculative trajectory on a physical robot can cause unsafe behavior. Traditional cache verification based on exact state matching is unsuitable in continuous environments due to friction, actuation noise, and sensor variance.

SPO therefore employs a continuous-state Local Edge Verifier. At each control step, the edge node observes the current state $s_t$ and compares it with the cached predicted state $\hat{s}_t$ using a normalized distance metric,
$e_t = \sqrt{(s_t - \hat{s}_t)^T W (s_t - \hat{s}_t)}$,
where $W$ is a diagonal normalization matrix with elements defined as the inverse variance of each state component estimated from offline calibration data.

Safety is enforced via a deterministic $\epsilon$-tube around the predicted trajectory in this normalized state space. Because verification occurs at every control step, deviations are detected within a single control cycle. If $e_t \le \epsilon_{base}$ (line 9--11), a Cache Hit is registered and the cached action $a_t$ is executed immediately. Otherwise, a Cache Miss occurs, the edge node invalidates the cached trajectory, issues a zero-velocity \emph{hold} command to safely stop the robot, and requests synchronous replanning from the cloud (line 12--16).

\begin{algorithm}[!htbp]
\caption{Speculative Policy Orchestration (SPO)}
\label{alg:spo}
\begin{algorithmic}[1]
\Require Action Policy $\pi_\theta$, World Model $\mathcal{M}_\phi$
\Require Tolerance $\epsilon_{base}$, normalization matrix $W$, limits $K_{min}, K_{max}$, expansion rate $\beta$
\State Initialize Edge Cache $\mathcal{C} \leftarrow \emptyset$
\State Initialize horizon $K \leftarrow K_{min}$
\State Initialize violation error $e_{miss} \leftarrow 0$

\For{each control step $t = 0, 1, 2, \dots$}
    \State Observe actual physical state $s_t$
    
    \Statex \textcolor{blue}{// 1. Edge-Side Verification and Execution}
    \If{$\mathcal{C}$ is not empty}
        \State Pop speculative tuple $(\hat{s}_t, \hat{a}_t)$ from head of $\mathcal{C}$
        \State $e_t \leftarrow \sqrt{(s_t-\hat{s}_t)^T W (s_t-\hat{s}_t)}$
        
        \If{$e_t \leq \epsilon_{base}$}
            \State Execute $a_t \leftarrow \hat{a}_t$ \Comment{Cache Hit}
            \State \textbf{continue} to next step $t$
        \Else
            \State $e_{miss} \leftarrow e_t$
            \State Flush Cache: $\mathcal{C} \leftarrow \emptyset$ \Comment{Safety Violation}
            \State Execute $a_t \leftarrow \text{SafeStop}(s_t)$ \Comment{Signal Tier-3 compliant deceleration}
        \EndIf
    \Else
        \State Execute $a_t \leftarrow \text{SafeStop}(s_t)$ \Comment{Hold via local impedance control}
    \EndIf
    
    \Statex \textcolor{blue}{// 2. Cloud-Side Speculative Generation \& AHS}
    \If{$e_{miss} > 0$} 
        \State Danger ratio: $\rho \leftarrow e_{miss} / \epsilon_{base}$
        \State $K \leftarrow \max(K_{min}, \lfloor K / \rho \rfloor)$
    \Else 
        \State $K \leftarrow \min(K_{max}, K + \beta)$
    \EndIf
    \State Reset error: $e_{miss} \leftarrow 0$
    
    \State Initialize speculative state $\hat{s}_{temp} \leftarrow s_t$
    \For{$k = 1$ to $K$}
        \State Predict speculative action: $\hat{a}_k \leftarrow \pi_\theta(\hat{s}_k)$
        \State Predict next state: $\hat{s}_{k+1} \leftarrow \mathcal{M}_\phi(\hat{s}_k, \hat{a}_k)$
        \State Append $(\hat{s}_{k+1}, \hat{a}_k)$ to $\mathcal{C}$
    \EndFor
\EndFor
\end{algorithmic}
\end{algorithm}

In summary, SPO jointly addresses execution safety and efficiency by enforcing bounded tracking error via an edge-based $\epsilon$-tube and by adaptively scaling the speculative horizon according to real-time physical uncertainty.


\section{EXPERIMENTS}
\label{experiment}
Our experiments address four questions: i) Does SPO reduce idle and execution time under high network latency?; ii) Can the $\epsilon$-tube verifier prevent error accumulation during contact-rich interactions?; iii) Does AHS balance bandwidth efficiency and responsiveness across tasks?; and iv) How does performance vary with world model accuracy?





\begin{figure}[t]
      \centering
      
      \includegraphics[width=\linewidth]{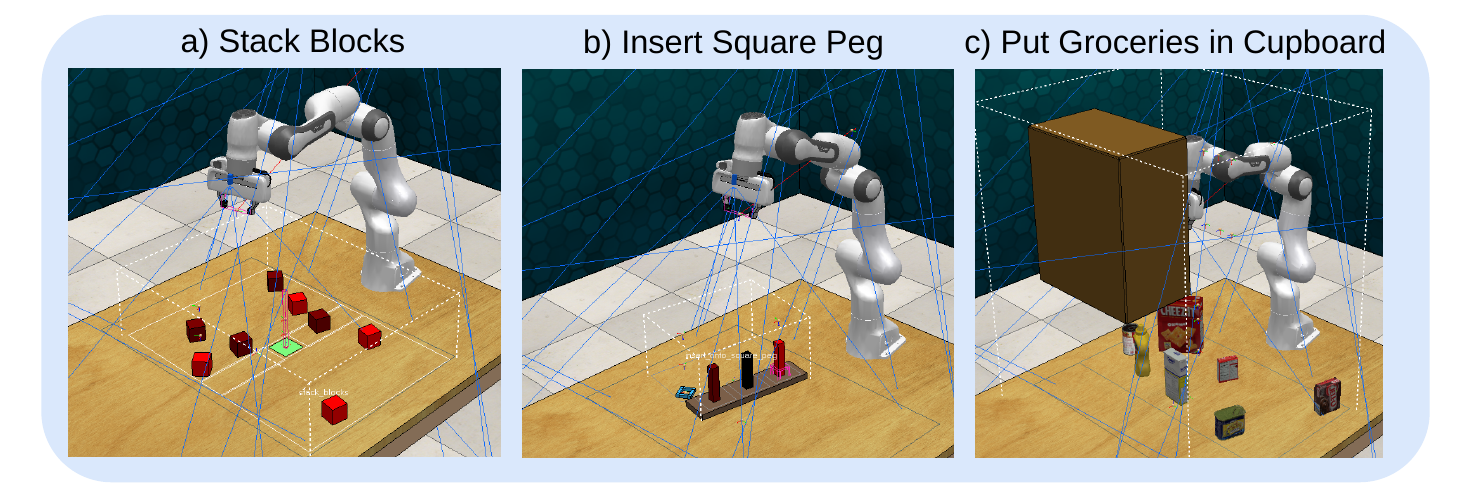}
    \caption{RLBench tasks used in the evaluation. From left to right, the tasks represent increasing levels of manipulation complexity.}
      \label{fig:rlbench_tasks}
   \end{figure}
     
\begin{figure}[t]
      \centering
      
      \includegraphics[width=\linewidth]{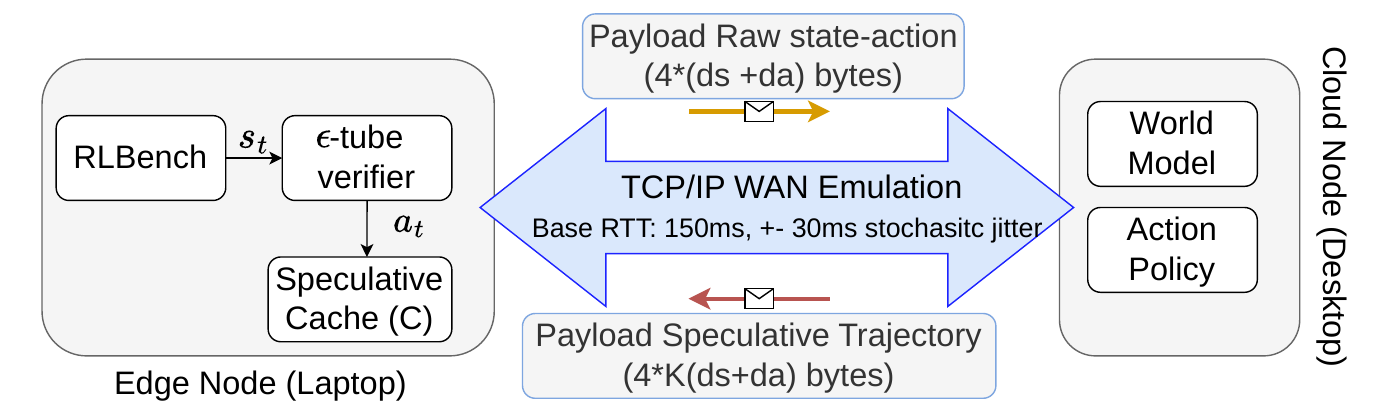}
  \caption{Distributed experimental setup where the edge runs RLBench and verification, while the cloud hosts the world model and policy.}
      \label{fig:experiment_setup}
   \end{figure}

\begin{table}[t]
\centering
\small
\setlength{\tabcolsep}{4pt}
\caption{Experimental Configuration}
\label{tab:parameters}
\begin{tabular}{lc}
\toprule
\textbf{Parameter} & \textbf{Value} \\
\midrule
\multicolumn{2}{l}{\textit{\textbf{Network}}} \\
Base round-trip delay ($\tau_{\text{net}}$) & 150 ms \\
Stochastic WAN jitter ($\tau_{\text{jitter}}$) & $\pm 30$ ms \\
Control frequency ($f_{ctrl}$) & 50 Hz \\
\midrule
\multicolumn{2}{l}{\textit{\textbf{SPO Horizon}}} \\
Min depth ($K_{min}$) & 2 \\
Max depth ($K_{max}$) & 10 \\
Additive horizon increment ($\beta$) & 1 \\
\midrule
\multicolumn{2}{l}{\textit{\textbf{Edge Safety}}} \\
Distance metric & Inverse-Variance Weighted \\
$\epsilon$-tube tolerance ($\epsilon_{base}$) & 20.0 \\
State dimension ($d_s$) & 141--295 \\
\midrule
\multicolumn{2}{l}{\textit{\textbf{Neural Models}}} \\
Architecture & 3-layer MLP \\
Hidden units & 512 \\
Action dimension ($d_a$) & 8 (7 joints + gripper) \\
\bottomrule
\end{tabular}
\end{table}

\subsection{Experimental Setup}
\subsubsection{Task Environments and MDP Configuration}

We evaluate SPO on three manipulation tasks from RLBench~\cite{james2020rlbench}. Each task is modeled as a continuous MDP with state space $\mathcal{S}$ containing robot proprioception (joint positions and velocities) and object poses, and an action space $\mathcal{A}$ defined as an 8D vector of 7-DOF joint velocities plus a discrete gripper command.
The tasks cover increasing levels of contact complexity (see Fig.~\ref{fig:rlbench_tasks}):

\begin{itemize}
    \item \textbf{Stack Blocks (Low Complexity, $d_s=148$):} Grasp a block and place it on a base block. This task mostly involves free-space motion and serves as a baseline for speculative execution.

    \item \textbf{Insert Onto Square Peg (Medium Complexity, $d_s=141$):} Insert a square ring onto a peg with tight geometric tolerances, testing the $\epsilon$-tube verifier’s ability to detect small alignment errors.

    \item \textbf{Put All Groceries In Cupboard (High Complexity, $d_s=295$):} A multi-stage pick-and-place task with multiple objects and occlusions, stressing adaptive horizon scaling under complex interactions.
\end{itemize}

\subsubsection{World Model and Policy}

\paragraph{Oracle Policy and World Model}
To isolate systems-level effects such as network latency, speculative caching, and verification, we instantiate the remote policy $\pi$ and world model $\mathcal{W}$ using an algorithmic Oracle derived from RLBench expert demonstrations. Given a state $s_t$, the Oracle policy $\pi^*$ returns the optimal action sequence $\mathbf{a}_{t:t+K}$, while the Oracle world model $\mathcal{W}^*$ provides the corresponding ground-truth rollout $\mathbf{\hat{s}}_{t+1:t+K}$. This perfect predictor ensures that any divergence detected by the $\epsilon$-tube verifier results from execution drift rather than model error, enabling direct evaluation of AHS under network latency $\tau_{\text{net}}$.

\paragraph{Learned Model Validation}
To assess robustness under imperfect prediction, we additionally train a lightweight three-layer MLP on RLBench demonstrations to autoregressively predict future states and actions. The resulting prediction drift allows us to evaluate how the verifier detects hallucinated rollouts and triggers safe cache invalidation.

\subsubsection{Hardware and Distributed Network Setup}

We deploy SPO across two TCP/IP-connected machines: an Edge Node running RLBench and the local verifier, and a Cloud Node hosting the neural models. To emulate WAN conditions, we inject a round-trip latency of $\tau_{\text{net}} = 150$\,ms with $\pm 30$\,ms jitter. Figure~\ref{fig:experiment_setup} shows the setup.

Communication uses ZeroMQ\footnote{\url{https://zeromq.org/}} with an uncompressed byte-array format. Each predicted state-action pair contains $d_s + d_a$ floats, requiring $4(d_s+d_a)$ bytes. For example, in \textit{Stack Blocks} ($d_s=148$, $d_a=8$), each speculative step requires 624 bytes, yielding $624 \times K_{max}$ bytes per cached trajectory.

\subsubsection{Baselines for Comparison}

We compare SPO against three baseline control strategies representing fixed-horizon caching:

\begin{itemize}
    \item \textbf{Synchronous Remote Inference (Blocking, $K=0$):} 
    The edge node sends state $s_t$ to the cloud and waits for a single action before continuing. No trajectory cache is used ($\mathcal{C}=\emptyset$), so each step incurs the full network latency $\tau_{\text{net}}$.

    \item \textbf{Top-1 Speculative Caching (T1-SC, $K=1$):} 
    A minimal speculative strategy that pre-fetches one state–action pair for the next step.

    \item \textbf{Naive Full-Tree Caching (NFTC, $K=10$):} 
    A static strategy that predicts a fixed trajectory of length $K=10$. Longer open-loop horizons can accumulate prediction error, particularly in contact-rich phases~\cite{bharadhwaj2024roboagent}.
\end{itemize}

\subsubsection{Performance Metrics}

We evaluate SPO and the baselines using the following metrics:

\begin{itemize}
\item \textbf{Task Success Rate (\%):} Percentage of trials where the task completes before the maximum allowed number of control steps ($\texttt{max\_steps}$).

\item \textbf{Cumulative Idle Time (s):} Total time the robot remains stationary while waiting for network responses or cloud inference.

\item \textbf{Cache Hit Rate ($\eta$):} Fraction of control steps served from the speculative cache versus requiring cloud fallback.

\item \textbf{Mean Horizon Depth ($\bar{K}$):} Average speculative horizon during execution, reflecting the AHS controller's adaptation to task dynamics.
\end{itemize}

Unless otherwise specified, the control interval is 20\,ms and the speculative horizon is bounded by $K_{min}=2$ and $K_{max}=10$. Additional parameters are listed in Table~\ref{tab:parameters}.

\subsection{Results and Discussion}

\begin{table}[h]
\centering
\caption{Quantitative Results with Oracle models. Results represent the theoretical performance upper bound using ground-truth predictors under $\tau_{\text{net}} = 150 \pm 30$ ms latency.}
\label{tab:oracle_results}
\resizebox{\columnwidth}{!}{
\begin{tabular}{ll cccc}
\toprule
\textbf{Task} & \textbf{Method} & \textbf{SR (\%)} & \textbf{Steps} & \textbf{Wall-time (s)} & \textbf{Idle time (s)} \\
\midrule
\multirow{4}{*}{\rotatebox[origin=c]{90}{Stack}}
& Blocking & 0 & 600  & 17.3 $\pm$ 0.1s & 10.7 $\pm$ 0.0 \\
& T1-SC  & 0 & 600 & 17.3 $\pm$ 0.1 & 9.7 $\pm$ 0.1 \\
& NFTC  & 100 & 506.0 $\pm$ 6.0 & 15.4 $\pm$ 0.1s & 4.5 $\pm$ 0.1 \\
& SPO  & 100 & 518.8 $\pm$ 3.7 & 15.7 $\pm$ 0.1 & 4.8 $\pm$ 0.1 \\
\midrule
\multirow{4}{*}{\rotatebox[origin=c]{90}{Insert}}
& Blocking  & 0 & 600 & 17.3 $\pm$ 0.0 & 10.7 $\pm$ 0.0 \\
& T1-SC  & 0 & 600 & 17.3 $\pm$ 0.0 & 9.7 $\pm$ 0.0 \\
& NFTC  & 100 & 280.4 $\pm$ 1.7 & 10.8 $\pm$ 0.0 & 2.5 $\pm$ 0.0 \\
& SPO  & 100 & 291.4 $\pm$ 5.2 & 11.0 $\pm$ 0.1 & 2.7 $\pm$ 0.1 \\
\midrule
\multirow{4}{*}{\rotatebox[origin=c]{90}{Put}}
& Blocking & 0 & 3000 & 65.9 $\pm$ 0.1 & 53.6 $\pm$ 0.0 \\
& T1-SC & 0 (1/7) & 3000 & 66.3 $\pm$ 0.0 & 48.4 $\pm$ 0.1 \\
& NFTC & 0 (6/7) & 3000 & 65.8 $\pm$ 0.0 & 56.5 $\pm$ 0.0 \\
& SPO & 0 (6/7) & 3000 & 65.9 $\pm$ 0.0 & 56.5 $\pm$ 0.0 \\

\bottomrule
\end{tabular}%
}
\end{table}

\begin{figure*}[t]
\centering
\begin{subfigure}[t]{0.3\textwidth}
\centering
\includegraphics[width=\linewidth]{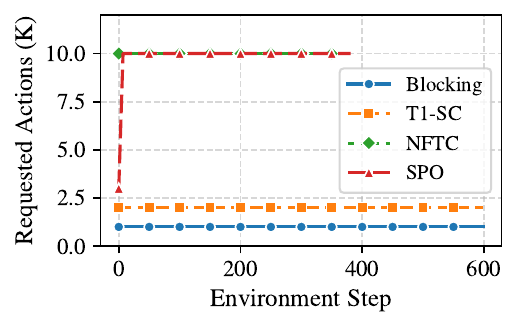}
\caption{StackBlocks}
\end{subfigure}
\hspace{1mm}
\begin{subfigure}[t]{0.3\textwidth}
\centering
\includegraphics[width=\linewidth]{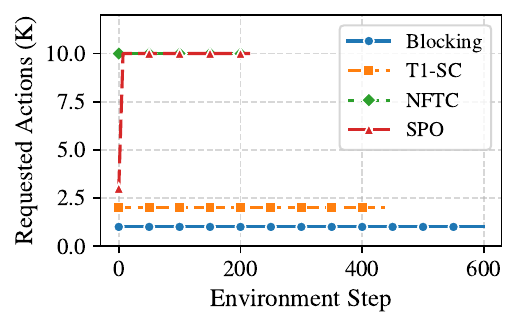}
\caption{InsertOntoSquarePeg}
\end{subfigure}
\hspace{1mm}
\begin{subfigure}[t]{0.3\textwidth}
\centering
\includegraphics[width=\linewidth]{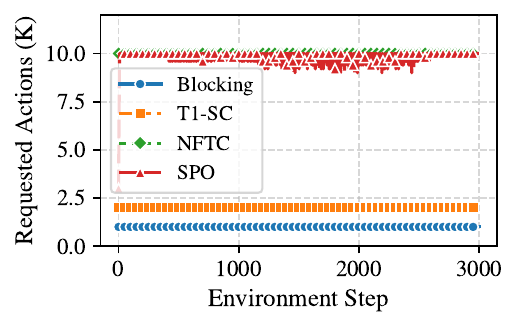}
\caption{PutAllGroceriesInCupboard}
\end{subfigure}
\caption{Adaptive Horizon Scaling Dynamics.}
\label{fig:ahs}
\end{figure*}

\begin{figure*}[t]
\centering
\begin{subfigure}[t]{0.3\textwidth}
\centering
\includegraphics[width=\linewidth]{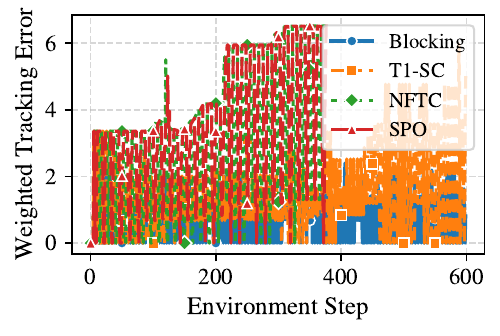}
\caption{StackBlocks}
\end{subfigure}
\hspace{1mm}
\begin{subfigure}[t]{0.3\textwidth}
\centering
\includegraphics[width=\linewidth]{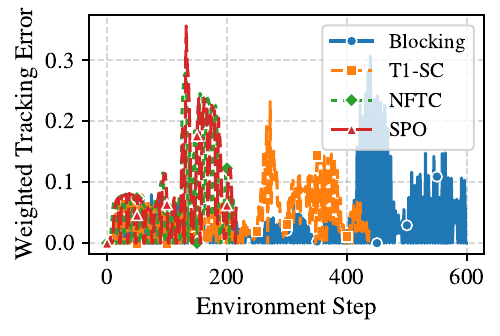}
\caption{InsertOntoSquarePeg}
\end{subfigure}
\hspace{1mm}
\begin{subfigure}[t]{0.3\textwidth}
\centering
\includegraphics[width=\linewidth]{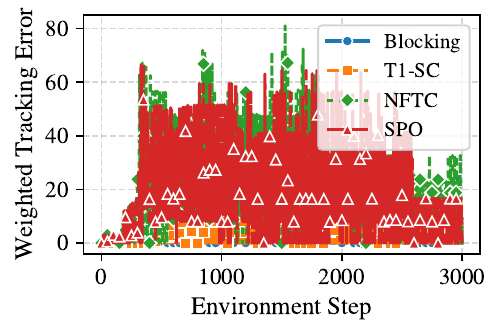}
\caption{PutAllGroceriesInCupboard}
\end{subfigure}
\caption{Tracking Fidelity and Safety Verification.}
\label{fig:safety_verification}
\end{figure*}
\subsubsection{Experiment Evaluation with Oracle Models}
We first evaluate SPO using Oracle predictors to establish a theoretical performance upper bound under high latency ($150 \pm 30$ ms). As shown in Table \ref{tab:oracle_results}, SPO effectively bridges the network-control gap that renders synchronous baselines (Blocking) non-functional.

\noindent\textbf{Efficiency and Task Success.} In StackBlocks and InsertSquarePeg, SPO achieves a 100\% success rate, whereas Blocking and T1-SC baselines fail due to persistent network stalls. Both SPO and NFTC saturate the speculative horizon at $K=10$ (200 ms of temporal autonomy), achieving near-identical wall-clock times (~15.7s for StackBlocks). This confirms SPO matches the theoretical speed of a static full-tree approach. In the complex PutAllGroceries task, while global success is 0\% due to fixed-path fragility, SPO and NFTC both achieve high sub-goal completion (6/7 items), vastly outperforming synchronous methods.

\noindent \textbf{Adaptive Scaling and Safety.} The 10-step buffer ($K=10$) at 50 Hz provides the necessary buffer to bridge the 150 ms RTT (Fig. \ref{fig:ahs}). While NFTC remains statically pinned at $K=10$, SPO’s AHS logic autonomously identifies this optimal depth while maintaining closed-loop verification.

The divergence between methods is most evident in tracking fidelity (Fig. \ref{fig:safety_verification}). During contact-rich phases in the PutAllGroceriesInCupboard task, NFTC exhibits significant error spikes, indicating that its open-loop execution cannot account for physical drift. Conversely, SPO utilizes a weighted verifier ($W$) to cap errors at the $\epsilon_{base}$ threshold. By triggering safety flushes, SPO maintains the physical fidelity required for precise manipulation, whereas Blocking and T1-SC baselines fail to progress, showing high residual error long after SPO has finished.

\subsubsection{Experimental Evaluation with Learned Models}
We train both the policy and world models using expert demonstrations collected in RLBench for the
\textit{InsertOntoSquarePeg} task. Each demonstration trajectory is processed to construct two datasets: (i) state--action pairs used for
policy learning, and (ii) state trajectories used to train the world model that captures the task dynamics. 

\begin{figure*}[t]
\centering
\begin{subfigure}[t]{0.23\textwidth}
\centering
\includegraphics[width=\linewidth]{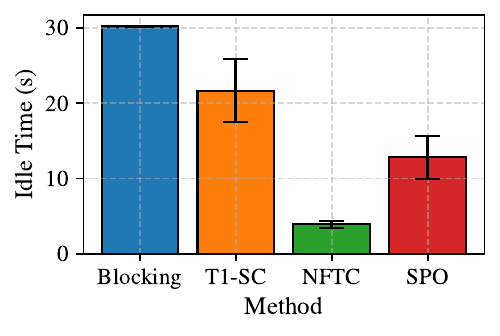}
\caption{Network idle time}
\label{fig:idle_time}
\end{subfigure}
\hspace{1mm}
\begin{subfigure}[t]{0.23\textwidth}
\centering
\includegraphics[width=\linewidth]{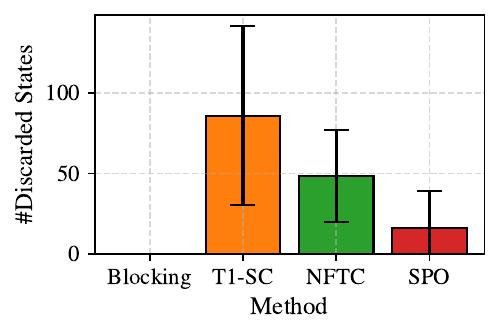}
\caption{Wasted cloud predictions}
\label{fig:wasted_predictions}
\end{subfigure}
\begin{subfigure}[t]{0.23\textwidth}
\centering
\includegraphics[width=\linewidth]{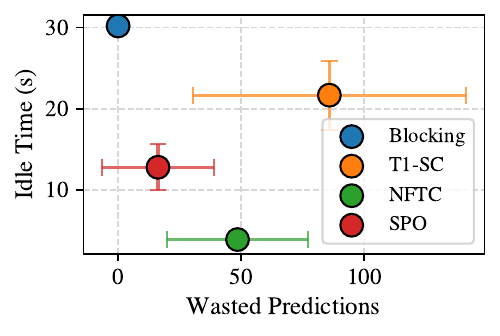}
\caption{Efficiency trade-off}
\label{fig:tradeoff}
\end{subfigure}
\hspace{1mm}

\caption{Quantitative results for \textit{InsertOntoSquarePeg} using learned models.}
\label{fig:learnt}

\end{figure*}

As shown in Figure~\ref{fig:learnt} for the \textit{InsertOntoSquarePeg} task, SPO achieves near-optimal latency masking, reducing network idle time by over 60\% compared to the Blocking baseline (Fig.~\ref{fig:idle_time}). Crucially, SPO achieves this efficiency while discarding approximately 60\% fewer cloud predictions than the NFTC ($K=10$) baseline (Fig.~\ref{fig:wasted_predictions}). The efficiency trade-off plot (Fig.~\ref{fig:tradeoff} ) confirms that SPO is the only method that minimizes both idle time and computational waste, proving that adaptive speculation is a resource-efficient necessity for cloud-native manipulation. It is worth noting that while our learned policy and world models achieve only modest accuracy, the SPO framework remains robust.

\subsubsection{Deployment Guidelines}

Deploying SPO in a new environment requires configuring three parameters: $K_{max}$, $K_{min}$, and $\epsilon_{\text{base}}$.
The maximum speculative horizon $K_{max}$ is selected based on the target network latency $\tau_{\text{net}}$ and the control interval $\Delta t$, such that
$K_{max} \cdot \Delta t \ge \tau_{\text{net}}$, ensuring sufficient buffering to mask network delay.
The minimum horizon $K_{min}$ is set to a small constant (e.g., 1–2 steps) to preserve a minimal local buffer during disturbance-prone or contact-rich phases.
Finally, the safety tolerance $\epsilon_{\text{base}}$ is determined by task geometry and allowable tracking deviation. Tasks with tight clearances (e.g., insertion) require smaller tolerances, whereas free-space motions may admit larger bounds to improve cache utilization.

\noindent\textbf{Remark on Hardware Deployment:} Algorithm~\ref{alg:spo} uses a strict zero-velocity hold ($a_t = \mathbf{0}$) to cleanly isolate network-induced idle time in simulation. However, commanding instantaneous zero velocity on physical robots causes severe mechanical stress and stick-slip artifacts during contact. In real-world deployments, this algorithmic reflex must map to a Tier-3 hardware safe stop (e.g., jerk-limited deceleration with impedance control) to safely dissipate kinetic energy.

\section{CONCLUSIONS}

In this paper, we presented Speculative Policy Orchestration (SPO), a cloud--edge framework for latency-resilient continuous robotic control. By combining speculative trajectory rollouts in the cloud with an edge-based $\epsilon$-tube verifier, SPO decouples high-frequency physical execution from network round-trip delay while maintaining bounded kinematic deviation.

Across multiple RLBench manipulation tasks spanning free-space and contact-rich scenarios, SPO reduced cumulative idle time relative to blocking remote inference and static fixed-horizon caching strategies. The Adaptive Horizon Scaling mechanism enabled dynamic adjustment of speculative depth, balancing latency masking and bandwidth efficiency under varying physical uncertainty.
Overall, SPO demonstrates that speculative orchestration can reconcile cloud-scale models with real-time robotic control constraints.

Future work will investigate deployment on physical robotic platforms to evaluate robustness under sensor noise and sim-to-real dynamics gaps. We also plan to integrate larger-scale vision-language-action models within the SPO framework to study its applicability to foundation-model-driven manipulation systems.








\bibliographystyle{IEEEtran}
\bibliography{acmart}

\end{document}